\documentclass[preprint]{elsarticle}

\usepackage{lineno,hyperref}
\usepackage{times}
\usepackage{epsfig}
\usepackage{graphicx}
\usepackage{amsmath}
\usepackage{amssymb}
\usepackage{subfigure}
\usepackage{epstopdf}
\usepackage{color}
 \usepackage{multirow}

\journal{Neurocomputing}

\begin{document}

\begin{frontmatter}

\title{A Fast Face Detection Method via Convolutional Neural Network}

\author[mymainaddress]{Guanjun Guo}
\ead{gjguo@stu.xmu.edu.cn}

\author[mymainaddress]{Hanzi Wang\corref{mycorrespondingauthor}}
\ead{hanzi.wang@xmu.edu.cn}

\author[mymainaddress]{Yan Yan}
\address[mymainaddress]{Fujian Key Laboratory of Sensing and Computing for Smart City, and School of Information Science and Engineering, Xiamen University, Xiamen 361005, P. R. China.}
\ead{yanyan@xmu.edu.cn}

\author[mysecondaryaddress]{Jin Zheng}
\ead{jinzheng@buaa.edu.cn}
\author[mysecondaryaddress]{Bo Li}
\cortext[mycorrespondingauthor]{Corresponding author.}
\ead{boli@buaa.edu.cn}
\address[mysecondaryaddress]{Beijing Key Laboratory of Digital Media, and School of Computer Science and Engineering, Beihang University, Beijing 100191, P. R. China.}

\begin{abstract}
Current face or object detection methods via convolutional neural network (such as OverFeat, R-CNN and DenseNet) explicitly extract multi-scale features based on an image pyramid. However, such a strategy increases the computational burden for face detection. In this paper, we propose a fast face detection method based on discriminative complete features (DCFs) extracted by an elaborately designed convolutional neural network, where face detection is directly performed on the complete feature maps. DCFs have shown the ability of scale invariance, which is beneficial for face detection with high speed and promising performance. Therefore, extracting multi-scale features on an image pyramid employed in the conventional methods is not required in the proposed method, which can greatly improve its efficiency for face detection. Experimental results on several popular face detection datasets show the efficiency and the effectiveness of the proposed method for face detection.
\end{abstract}


\end{frontmatter}


\section{Introduction}
In the Marr's theory of vision \cite{MarrVision}, a representational framework for vision was proposed, which consists of three levels: primal sketch, 1.5-D sketch and 3-D model representation. How to extract a proper image representation corresponding to the three levels is a critical problem in computer vision. So far, there are some milestone representations during the development of computer vision. For example, SIFT \cite{sift2004} and HOG \cite{hog2005} features have exhibited the good property of local invariance. Canny features \cite{Canny1986} can capture the 1.5-D sketch, which represents the low-level image structure. The deconvolutional network \cite{deconv11} tries to characterize both mid-level and high-level representations in an unsupervised manner to understand images. However, finding a high-level representation for different computer vision tasks is still challenging.

Deep learning models have demonstrated impressive performance for different computer vision applications \cite{girshick14CVPR,sppnet2014,imageNet14,TAO201698,TIVIVE20083253,Lu201731,Guo2018TMM}.
The deep convolutional neural network (CNN) \cite{yannCNN} can map raw data from a manifold to the Euclidean space, in which features may be linearly separable.  Generally speaking, there are two ways to extract features with deep CNN models. The first way is to extract features on an image pyramid obtained by using object proposals generation methods (such as SelectiveSearch~\cite{SelectiveSearch13}, Edgeboxes~\cite{edgeBoxes2014} and CM~\cite{guo2017}) or the sliding-window strategy (such as R-CNN \cite{girshick14CVPR} and DenseNet \cite{denseNet}). However, this way suffers from high computational burden. The second way is to obtain features by using the sliding-window strategy on convolutional feature maps (such as OverFeat \cite{overfeat13} and faster R-CNN~\cite{fasterRCNN}). The second way is usually more efficient to obtain features than the first way, when the sliding-window moves at each position. However, one problem for the second way is that the obtained features are less discriminative than those obtained by the first way.
In addition, the computational complexity of both ways is high for object detection. Therefore, how to improve the efficiency of object detection remains a challenging problem due to the real-time requirement of object detection.

Face detection is a special object detection task, and it can be tackled by state-of-the-art object detection methods. For example, R-CNN, faster R-CNN and YOLO have been respectively extended to the task of face detection~\cite{jiang2016,Triantafyllidou2017,jiang2016,SunWH17,yolo2016}. However, as mentioned above, these object detection methods are usually time-consuming. Moreover, these methods mainly rely on the input generic object proposals obtained by an object proposal generation method. Since the obtained object proposals may not cover small-sized faces,  these detection methods are usually not effective at detecting small-sized faces.

\begin{figure}[t]
\begin{center}
   \includegraphics[width=0.8\linewidth]{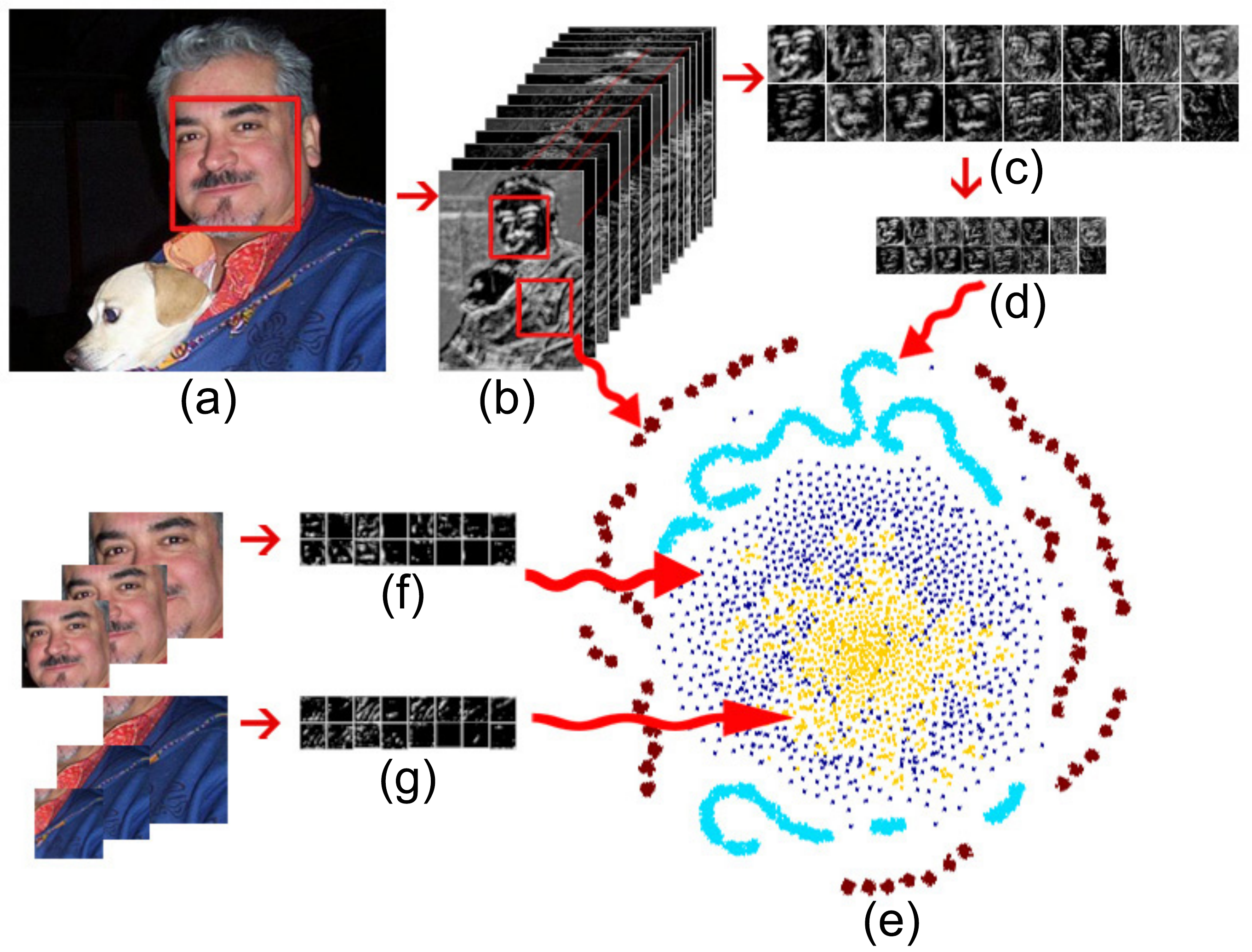}
\end{center}
   \caption{The visualization of DCFs and general CNN features by using the t-SNE \cite{maaten2008visualizing} visualization algorithm. (a) A test image. (b) The corresponding DCFs of the test image. (c) The features obtained from a set of the sliding windows. (d) Resizing the obtained features into the size of the fully-connected layer of CNN. (e) The visualization of the distribution of DCFs (shown in the brown and cyan dots) and general features of CNN (shown in the blue and yellow dots). (f) and (g) The general features extracted based on an image pyramid via CNN.}
\label{framework}
\end{figure}
In this paper, we propose a fast face detection method based on Discriminative Complete Features (DCFs) extracted by an elaborately designed deep CNN. The extracted DCFs are sparse and insensitive to scale variations for face detection. As a result, direct classification on DCFs significantly improves the efficiency of face detection compared with several state-of-the-art face detection methods using CNN.
The three key components of the proposed fast face detection method are summarized as follows: First, we generate a sparse feature space, which is implemented by using the rectified linear unit (ReLU)~\cite{relu2010}. Second, the multi-scale features are resized to the desirable size (i.e., the same as the size of the input to a fully-connected layer) by using the nearest neighbor interpolation method~\cite{Gonzalez2006}, which can keep the sparseness and the topological structure of features. Third, for each window in an original input image, we propose a fast method to obtain the features based on the complete feature maps \cite{giustiICIP13}. Based on the above three components, the desired features (i.e., DCFs) obtained from the feature maps before the fully-connected layer are extracted. As shown in Fig.~\ref{framework}, DCFs are linearly separable and sparse. Besides, by considering the fully-connected layer in CNN as a linear classifier, the generalization error bound of the linear classifier can be further calculated for face detection.

As the major contribution of this paper, a fast face detection method based on CNN is proposed. The sliding-window strategy can be used on the proposed discriminative complete features, which reduces the computational complexity of the proposed face detection method. Furthermore, the convolution operation in CNN is speeded up by using the Sparse Fast Fourier Transform algorithm. Experiments show the effectiveness and efficiency of the proposed face detection method.

The remainder of this paper is organized as follows. In Section 2, we review related work on object detection and face detection based on CNN. In Section 3, we propose a fast DCFs-based method by using a deep CNN for face detection. In Section 4, we conduct experiments to show the effectiveness of the proposed method and its performance on face detection. The main conclusions are summarized in Section 5.

\section{Related Work}
From low-level vision representations to high-level features, the importance of feature invariance under different conditions (such as changes in illumination and scale) is obvious. DiCarlo et al. \cite{Neuron2012} review the evidences of neurons and computational models, and discover the computation process which can mimic powerful neuronal representations in the inferior temporal cortex. The authors point out that two neural pathways via a series of visual cortical areas try to untangle the neuronal representations for object recognition and object detection. These two neural pathways are widely known as the ventral pathway and the dorsal pathway, respectively. In some literatures~\cite{DNNVS2015,TwoStream2014}, CNN plays a role similar to the ventral pathway, where the hierarchical features of CNN correspond to the neuronal representations from the visual area 1 (V1) to the visual area 4 (V4).
CNN works well for many different vision tasks by deeply learning the features of data. For instance, Farabet et al. \cite{6338939} show the impressive performance of CNN on scene labeling. Alex et al. \cite{alexNet2012} and He et al. \cite{he15deepresidual} use CNN to classify the ImageNet dataset \cite{imageNet14} and achieve the accuracy of 83\% and 94.3\%, respectively. R-CNN~\cite{girshick14CVPR}  improves mean average precision (mAP) by more than 30\% relative to the previous best result on the VOC 2012 dataset~\cite{Everingham10},  and it achieves a mAP of 53.3\%.

Although CNN has shown good performance for object detection, its computational efficiency is not high. OverFeat \cite{overfeat13} obtains the multi-scale dense features by using the sliding window strategy, and it achieves mAP of 24.3\% in the ILSVRC2013 competition \cite{imageNet14}. However, the detection process of OverFeat is very time-consuming, because it requires running the classifier and regressor networks across all the possible locations and scales. To improve the efficiency, R-CNN first generates many class-independent proposal windows, and then extracts features on the proposal windows with the trained CNN on a multi-scale images pyramid. Afterwards, a score is assigned to each proposal window by applying a linear SVM to the features. R-CNN takes about 15 seconds to run a detector on a $500\times375$ image. Another recently proposed method, called SPP-net \cite{sppnet2014}, removes the constraint of the fixed-sized input in the conventional CNN by replacing the pooling layer with a spatial pyramid pooling layer. Instead of extracting features from image regions, SPP-net directly obtains features from the feature maps before the fully-connected layer. However, SPP-net can not deal with flexible CNN architectures for different vision tasks. To reduce the computational
complexity of the CNN-based object detection methods and design flexible CNN architectures, Giusti et al. \cite{giustiICIP13} fragment the extended maps generated from each max-pooling layer. When a sliding-window moves in a test image, the corresponding features are obtained by looking up all the fragments in each layer for classification. This object detection method is theoretically faster than the other object detection methods in which features are obtained by using a CNN model on each patch in an image. However, considering the image size and the number of layers, this method is still not computationally efficient. Based on the above reviews, we can see that most current object detection methods based on deep learning still consume much time during the object detection process.

Several object detection methods based on CNN are applied to the task of face detection. For example, Lin et al.~\cite{LIN2016} propose the MLetNet method, where the number of the output nodes of the LeNet CNN structure~\cite{yannCNN} is modified for face detection, to detect the masked faces of possible terrorists. FaceCraft~\cite{Qin2016} jointly trains a region proposal network (RPN)~\cite{fasterRCNN} and a fast R-CNN model~\cite{Girshick2015} to improve the detection rate of the fast R-CNN method for face detection. However, as mentioned above, fast R-CNN is time-consuming since it extracts features on an image pyramid. MTCNN~\cite{mtcnn2016} utilizes a cascaded multi-task framework, which exploits the inherent correlation between the tasks of detection and alignment, to boost up the performance of both face detection and alignment. Since MTCNN  uses multiple CNN networks to detect faces, it is still not computationally efficient. Several other face detection methods also apply fast R-CNN or faster R-CNN to the task of face detection, such as~\cite{Triantafyllidou2017,jiang2016}. However, these two methods are not effective at detecting small-sized faces since the object proposal generation methods are designed for generic objects instead of faces. Recently, SAFD~\cite{zekun17} uses a scale proposal network to predict the possible scales of the faces in a test image, which can effectively boost the performance of CNN based face detectors.

Next, we propose a fast face detection method by directly using the sliding window strategy on discriminative complete feature maps.
\section{A Fast Face Detection Method Based on Discriminative Complete Features}
In this section, we describe the details of the proposed fast face detection method via CNN. In Section 3.1, we give the overview of the proposed method.
In Section 3.2, we show the details of the proposed DCFs. In Section 3.3, a fast face detection method is given by performing direct classification on DCFs. In Section 3.4, an ensemble model is used to further reduce the classification error on DCFs.
\subsection{Overview of the Proposed Method}
\begin{figure*}[t]
\begin{center}
   \includegraphics[width=0.92\linewidth]{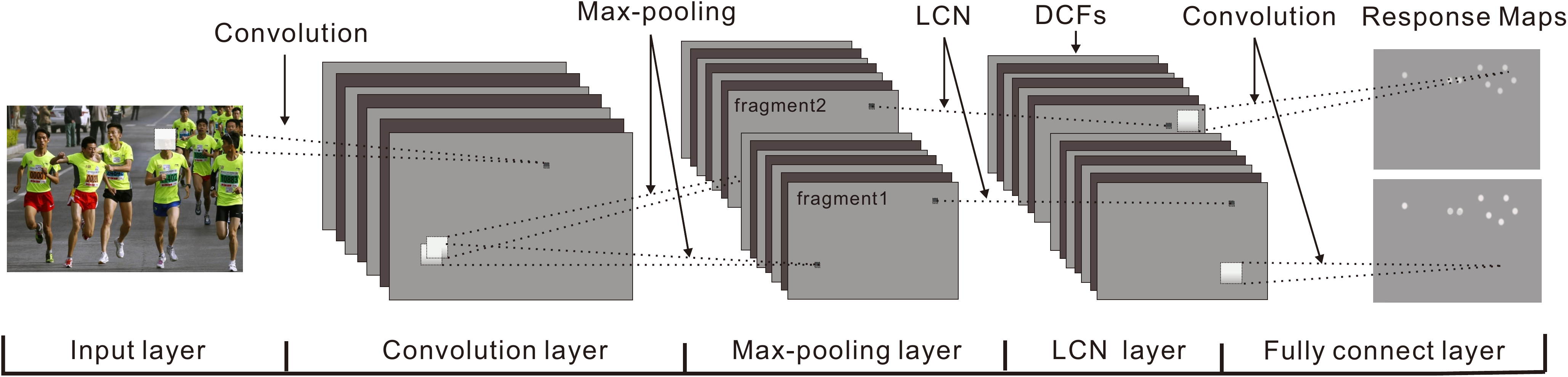}
\end{center}
   \caption{The proposed DCFs-based method for object detection. Different fragments are created corresponding to different offsets in a max-pooling layer, which preserves all the possible discriminative features. LCN is the abbreviation of Local Contrast Normalization.}
\label{DCFFramework}
\end{figure*}
In this paper, we propose a fast method for face detection by directly using the sliding window strategy on the feature maps before the fully-connected layer, where a specific CNN is elaborately designed, as shown in Fig.~\ref{DCFFramework}. We expect the trained CNN model can map the whole input image to a feature space, in which the new features in each sliding window are linearly separable. Then, the computational complexity of the proposed face detection method is significantly reduced by performing classification on the feature space. Since the classification layer in a CNN model is usually a linear classifier, we choose the output of the layer before the classification layer as the new features. All the layers before the classification layer act as a nonlinear mapping function (see Fig.~\ref{DCFFramework}).

The proposed method includes two key elements, which are the nonlinear mapping function and the sparse discriminative features for face detection. The nonlinear mapping function projects raw data onto the Euclidean space, and the sparseness of features is beneficial to linear separability. As shown in Fig.~\ref{DCFFramework}, the convolution layers and the max-pooling layers act as the nonlinear mapping function, where ReLU constrains the output features to be sparse. At the same time, the local contrast normalization (LCN) layers generate compact and competitive features.  The LCN layers are inspired by computational neuroscience models~\cite{Pinto_whyis}, where local competition between adjacent features in a feature map is enforced. It has been empirically proven that the LCN layers can reduce the error rates and  make supervised learning considerably faster~\cite{Jarrett2009}.
\subsection{Discriminative Complete Features (DCFs)}
 \begin{figure*}
\begin{center}
   \includegraphics[width=0.5\linewidth]{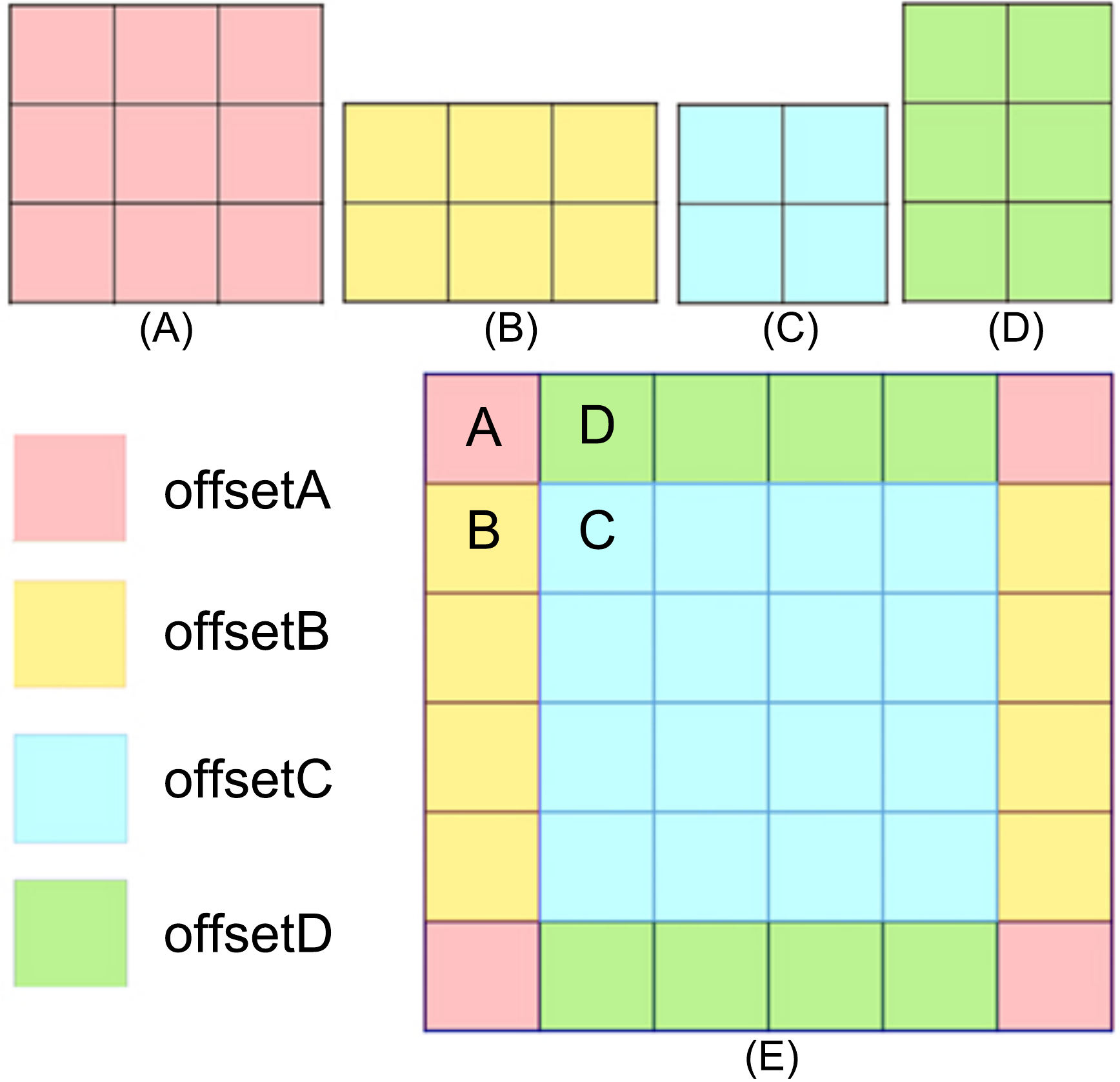}
\end{center}
   \caption{Illustration of the fragments of the extended feature maps based on different offsets of the max-pooling kernel. (A), (B), (C), (D) respectively denote different fragments of the extended maps after max-pooling corresponding to different offsets in (E). (E) denotes the input feature map for a max-pooling layer. For the kernel size of $2 \times 2$, A, B, C, D in (E) denote different offsets respectively.}
\label{Fragments}
\end{figure*}
To improve the efficiency of the proposed face detection method, we directly perform classification on the feature maps. However, in the process of forward propagation of CNN on a test image, one potential problem is that some discriminative information for face detection is lost when the operation of max-pooling is applied to the whole test image. As illustrated in Fig.~\ref{Fragments}, the max-pooling
layer partitions the input feature map (E) into a set of $2\times2$ non-overlapping kernels. For each kernel, it outputs the maximum pixel value. There are four ways to tile max-pooling kernels over the feature map (E) corresponding to the four different offsets in the max-pooling kernel. For convenience, the four different offsets are denoted as A, B, C, D, respectively. If the $2\times2$ non-overlapping max-pooling kernels are tiled over the input feature map (E) beginning with the offset $A$, the output feature map is shown in Fig.~\ref{Fragments}(A). For convenience, Fig.~\ref{Fragments}(A) is called as fragment (A). Similarly, fragments (B), (C) and (D) are obtained when the $2\times2$ non-overlapping max-pooling kernels are tiled over the input feature map (E) beginning with the offset B, C and D, respectively.  The general operation of the max-pooling step on the test image only obtains a fragment (A) after tiling non-overlapping max-pooling kernels over the feature map (E) starting with the offset A, and the other fragments with the other offsets are not obtained. If a sliding window moves at the other offsets (i.e., B, C, D), the obtained features may not be discriminative. Therefore, we adopt the strategy of~\cite{giustiICIP13} to keep all the possible fragments. The corresponding fragment is created when we tile non-overlapping max-pooling kernels over the feature map (E) in Fig.~\ref{Fragments} starting with different offsets (i.e., A, B, C, D), and each fragment is independent to the other fragments at the same layer. Based on the fragments in the extended feature maps, we can extract DCFs and directly perform classification on DCFs. The definition of each layer, which is used to extract DCFs, is given next.

 Given an input image $x$, the trained filter bank $k$ and bias $b$, the output of the convolution layer can be written as:
\begin{equation}\label{convEq}
  x_{oj}^l=f(\sum_{i\in{M_j}}{x_{oi}^{l-1}*k_{ij}^l+b_j^l}),
\end{equation}
where $o$ denotes the index of fragments; $l$ denotes the current layer; $i$ and $j$ denote the indexes of the input and output feature maps, respectively; $M_j$ represents a selection of the input feature maps, which are the output features from the previous layer;  $*$ denotes the operator of convolution; $f$ denotes the activation function, where $f(x)=max(0,x)$ is the ReLU \cite{relu2010} activation function. The output of ReLU is 0 if the input is less than 0, and raw output otherwise.  Thus, ReLU constrains the output features to be sparse.

For the max-pooling layer, we have
\begin{equation}\label{maxpoolEq}
  x_{oj}^l(m,n)=max(x_{oj}^{l-1}(p:(p+s),q:(q+s))),
\end{equation}
where $m$ and $n$ respectively denote the row and column index of a feature map in the current layer, respectively; $s$ denotes the kernel size of the max-pooling layer; $p=s\times(m-1)+\kappa+1$ and $q=s\times(n-1)+\kappa+1$, where $\kappa$ ($0<\kappa<s$) denotes the offset, $p$ and $q$ denote the starting positions of the row and column of each tiled max-pooling kernel, respectively; the colon is used to pick out the selected rows or columns of the feature maps, and here $o\leq s^2$. The max operation has the property of the local invariance~\cite{SchererEvaluation}.

Inspired by the computational neuroscience, the local contrast normalization (LCN) layer \cite{Pinto_whyis} mimics the cells in V1 to enforce local competition in the feature maps. In order to extract features from an input image with an arbitrary size, we set the local contrast normalization layer as:
\begin{equation}\label{lcnEq}
  \hat{x}_{oj}^l(m,n)=\frac{x_{oj}^l(m,n)}{(\kappa+\alpha\sum\limits_{i=max(0,j-\frac{r}{2})}^{min(N-1,j+\frac{r}{2})}{x_{oi}^l(m,n)^2})^\beta},
\end{equation}
where $r$ denotes the number of the adjacent feature maps (usually set to be a positive integer in the training process); $N$ denotes the total number of the feature maps at the current layer. $\kappa$, $\alpha$, $\beta$ are the hyperparameters which are set to be appropriate float point values in the training process. Eqs.~(\ref{convEq}), (\ref{maxpoolEq}) and (\ref{lcnEq}) can be used to extract the DCF features, where all the discriminative information for face detection is kept. Thus, we call the features obtained from the feature maps before the fully-connected layer as the discriminative complete features (DCFs). As a result, direct classification on the feature maps before the fully connected layer can improve the efficiency of the detection procedure.
\subsection{Speedup for Face Detection Based on DCFs}
Based on the above-mentioned three layers (i.e., the convolution layer, the max-pooling layer, the local contrast layer), the input images are mapped into a feature space, where the features for face detection are linear separable. Therefore, we can directly perform classification on DCFs extracted from each patch corresponding to a sliding window by using the weight vector of the fully-connected layer. Instead of using the sliding window technique on the input images, direct classification on DCFs can significantly improve the efficiency for face detection. As a matter of fact, the weight vector of the fully-connected layer can be reshaped into a kernel matrix so that the dot product between DCFs and the weight vector is transformed into the convolution between DCFs and the kernel matrix~\cite{fcn2017}. In addition, the convolution can be implemented by using the Sparse Fast Fourier Transform algorithm \cite{SFFT}, which also improves the speed for face detection since the DCFs of an image are sparse.

To show the efficiency of the proposed method for face detection, the theoretical analysis on the computational complexity of the floating point operations (FLOPS) required by the proposed method and a couple of other methods (including the patch-based method~\cite{girshick14CVPR} and the image-based method~\cite{giustiICIP13} ) is presented next. The patch-based method applies a trained CNN model to each overlapping patch in the test image. The image-based method performs the convolution only once for the test image. The main difference between the image-based method and the proposed DCFs-based method is that the image-based method applies the sliding window technique on the original test image, while the proposed DCFs-based method directly uses the sliding window technique on the feature maps. Since the size of the feature maps is far smaller than that of the original image, the proposed method is much more efficient.  In addition, the convolution operation in CNN is speeded up by using the Sparse Fast Fourier Transform algorithm in the proposed face detection method. The FLOPS required by the patch-based method $FLOPS_l^{patch}$, the image-based method $FLOPS_l^{Image}$ and the proposed DCFs-based method $FLOPS_l^{DCF}$ are given in detail next. Here, we only analyze the computational complexity of the FLOPS required by the three methods on the convolutional layer and the fully-connected layer, because these two layers are the most time-consuming.

For the convolutional layer, the FLOPS required by the patch-based method is written as:
\begin{equation}\label{PatchConvFLOPS}
\begin{aligned}
FLOPS_l^{patch}=2A|P_{l-1}||P_l|w_l^2s_l^2,
\end{aligned}
\end{equation}
where $A$, $w_l$, $s_l$ denote the number of the pixels of an input image, the number of the pixels of the feature map in the $l$th layer, and the kernel size at the $l$th layer, respectively; $|P_{l-1}|$ and $|P_l|$ denote the numbers of the feature maps in the previous layer and in the current layer, respectively. The patch-based method is quite time-consuming since the number of convolution operations on each feature map is equal to the number of pixels of an input image, which causes a large number of redundant computations.

In contrast, the FLOPS required by the image-based method is written as:
\begin{equation}\label{ImageConvFLOPS}
\begin{aligned}
FLOPS_l^{Image}=2A_l|P_{l-1}||P_l|F_ls_l^2,
\end{aligned}
\end{equation}
where $A_l$ denotes the number of the pixels in the feature map; $F_l$ denotes the number of the fragments in the current layer. As the size of feature maps decreases, the FLOPS required by the image-based method on the convolutional layer reduces.

The DCFs-based method utilizes the Sparse Fast Fourier Transform algorithm to improve the efficiency of convolution. Thus, the FLOPS required by the DCFs-based method can be written as:
\begin{equation}\label{OursConvFLOPS}
\begin{aligned}
FLOPS_l^{DCF}=2A_l^*\ln({A_l})|P_{l-1}||P_l|F_l,
\end{aligned}
\end{equation}
where $A_l^*$ denotes the number of the non-zeros pixels in the feature map.

\begin{figure}
\begin{center}
   \includegraphics[width=0.65\linewidth]{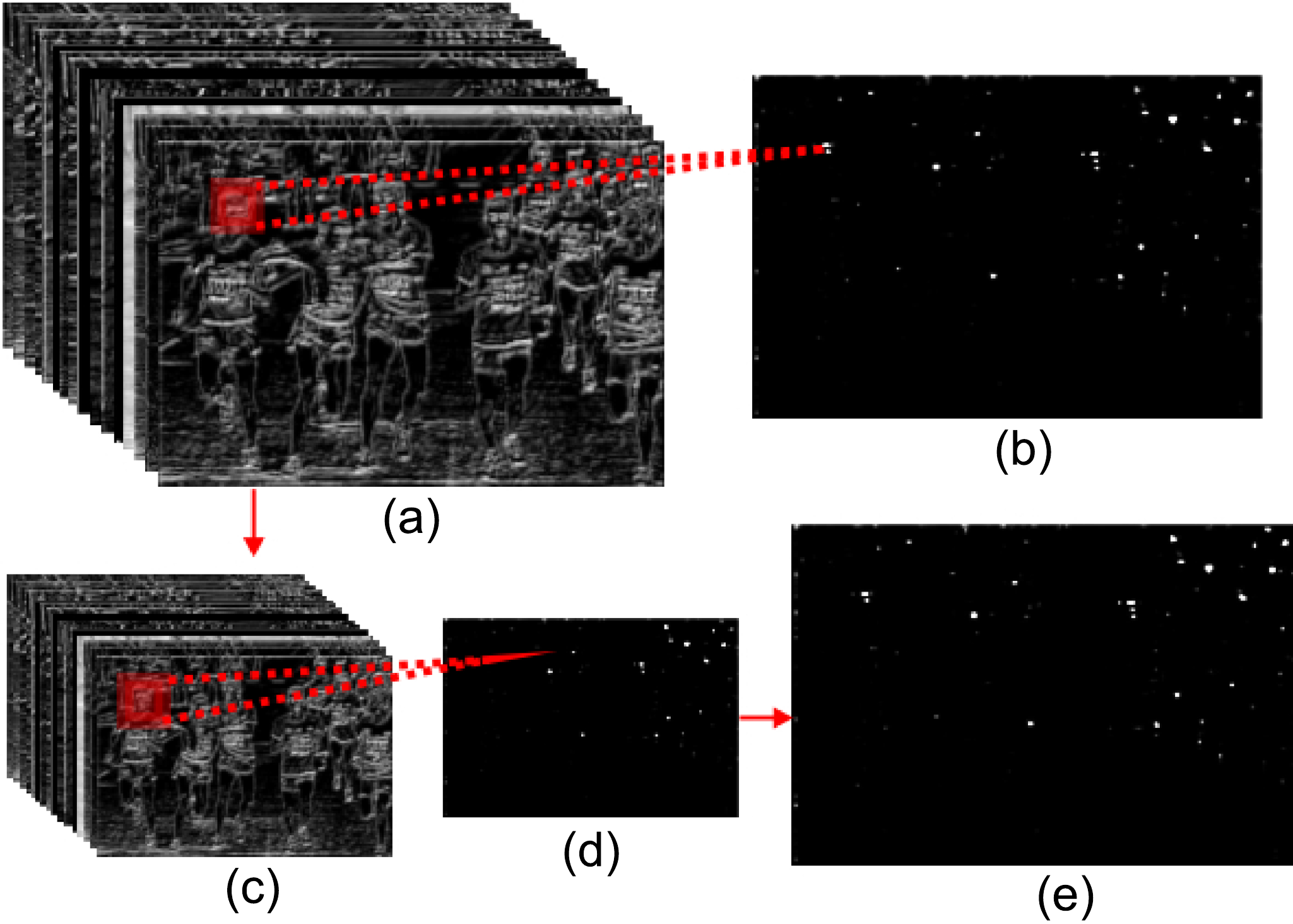}
\end{center}
   \caption{Face detection on the DCFs feature maps. (a) The DCFs obtained by the proposed method. (b) The response map corresponding to the DCFs at the last layer. (c) The resized feature maps obtained by the nearest neighbor interpolation method. (d) The output response map of the resized DCFs. (e) The resized response map (the same size as (b)) for further processing by using non-maximum suppression.}
\label{DetectionFig}
\end{figure}

For the fully-connected layer, the FLOPS required by the patch-based method, the image-based method and the DCFs-based method are respectively written as Eq.~(\ref{Eq::PatchflFLOPS}), Eq.~(\ref{Eq::ImageflFLOPS}) and Eq.~(\ref{Eq::OursflFLOPS}).
\begin{equation}\label{Eq::PatchflFLOPS}
\begin{aligned}
FLOPS_l^{patch}=2A|P_{l-1}||P_l|s_l.
\end{aligned}
\end{equation}
\begin{equation}\label{Eq::ImageflFLOPS}
\begin{aligned}
FLOPS_l^{Image}=2AF_ls_l.
\end{aligned}
\end{equation}
\begin{equation}\label{Eq::OursflFLOPS}
\begin{aligned}
FLOPS_l^{DCF}=2A_l^*\ln({A_l})2F_l.
\end{aligned}
\end{equation}
Since the proposed method uses the sliding window technique on DCFs instead of the original test image, the number of candidate windows used for classification in the proposed method is significantly reduced compared with that used in the other two methods (i.e., the patch-based method and the image-based method).

As the sliding window moves at each position of the feature map in a fragment of DCFs, a resulting response map is obtained by using the classifier layer on the feature vector obtained from each sliding window. Multi-scale DCFs, which are obtained by resizing DCFs with the nearest neighbor interpolation method, are also used for classification. The response maps of the multi-scale DCFs are shown in Fig. \ref{DetectionFig}.
The estimated scale of a face candidate is obtained by using the non-maximum suppression technique, and the response maps of all fragments are merged into one response map using non-maximum suppression. Finally, by backprojecting the estimated scale and the position of the centroids of the connected regions in the response map onto the original image, the locations of the faces can be detected with the estimated scale in the original image.
\subsection{An Ensemble Model for Reducing Generalization Error}
In this subsection, an ensemble model~\cite{emsemble2012} is used to improve the performance of the proposed method.
Based on the PAC theory \cite{Valiant1984}, which shows the theoretical analysis of a linear classifier, we can analyze the generalization error of the linear classifier at the fully-connected layer. In the framework of PAC, the PAC-bayesian margin bound is the upper bound of generalization error $R[w]$ (see \textbf{Theorem 3} in \cite{pac2001}). Then, the number of model instances can be computed. Specifically, given the upper bound of generalization error $R[w]$ and a generalization error threshold of the linear classifier $H[w]$, the number of model instances $v$ can be calculated as:
\begin{equation}\label{ensembleEq}
  v=\lceil{\frac{\ln(H[w])}{\ln(R[w])}}\rceil,
\end{equation}
where $\lceil . \rceil$ denotes the ceiling function.

\subsection{Face Bounding Box Regression}
Inspired by R-CNN~\cite{RCNN} and MDNet~\cite{mdnet2016}, we also train a regressor, which is used to regress the face bounding boxes. The regressor consists of two linear layers, where the first layer has one hundred output nodes and the second layer has four output nodes. The learning objective of the regressor is formulated as the mean squared error between the four parameter values of a candidate face window (i.e., the horizontal and vertical coordinate values of the top-left corner, the width and height of the face bounding box) and those of the ground truth face bounding box. For each candidate face window, its DCFs are used to train the regressor. We expect that the regressor can refine the face bounding box. Usually, the candidate face window is obtained by using the sliding window strategy on DCFs with several fixed scales. However, the resulting face bounding boxes may not be accurate enough, since the width and height of a face bounding box are continuous values instead of the values in several fixed scales. The regressor can overcome the above drawback by further refining the bounding box.
\section{Experiments}
In this section, the detailed evaluation of the performance of the proposed method on face detection is shown. Section 4.1 gives the parameter settings used in training the CNN model. Section 4.2 visualizes the trained CNN network. Section 4.3 presents the comparison of efficiency and performance between the proposed method and several state-of-the-art face detection methods on two public face datasets.

\subsection{Parameter Settings}
We use the GPU implementation of the  $cuda-convnet$ code \cite{alexNet2012} to train the CNN model on various datasets.
In order to adapt the parameters of the trained model to test images of arbitrary size, the batch normalization step in $cuda-convnet$ is not performed  during the training process.
\begin{table}\small
\begin{center}
\caption {\label{tbCNNStru}The CNN structure for face detection.}
\begin{tabular}{ | l | c | c | c| c| c| c | }
  \hline
  Layer  & Type & Input & \#Channels & \#Filters & Filter Size & Activation \\ \hline
  Layer1 & conv & Data  & 3         & 16       & 5$\times$5           &ReLU        \\ \hline
  Layer2 & pool & Layer1 & 16       &  -        &  -         &   -   \\ \hline
  Layer3 & cmrnorm & Layer2 & 16    &  -        &  -          &  -     \\ \hline
  Layer4 & conv & Layer3  & 16     & 16       & 5$\times$5           &ReLU        \\ \hline
  Layer5 & cmrnorm & Layer4 & 16    & -        &   -         &  -   \\ \hline
  Layer6 & pool & Layer5 & 16       & -        &  -          &  -     \\ \hline
  Layer7 & FC & Layer6 &    -   &  -       &     -       &    -  \\ \hline
  Layer8 & softmax & Layer7 &    -    &     -    &    -        &  -   \\
  \hline
\end{tabular}
\end{center}
\end{table}

\textbf{Data pre-processing:}
We collect 13,466 face images from the dataset provided by Sun et al. \cite{yisun2013}, 29,821 face images from the CAS-PEAL-R1 dataset \cite{Gao04thecas-peal} and 47,220 face images from the web. The collected face dataset totally includes 90,507 face images with different lighting conditions, poses and ages. Moreover, 100,000 non-face images from the VOC2012 dataset \cite{Everingham10} are collected for training. All the face and non-face images are converted to the gray images and resized to the size of 32$\times$32 . All images are split into 19 batches and 10,000 images per batch are used for training.

\begin{figure}
\begin{center}
   \includegraphics[width=0.8\linewidth]{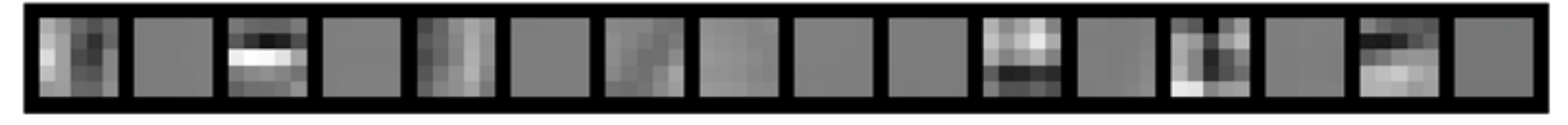}
   \caption{The trained filters of the first convolutional layer in the CNN model used by the proposed method for face detection.}
\label{convFilters}
\end{center}
\end{figure}

\textbf{Hyperparameters for training:} To balance the efficiency and effectiveness of the proposed face detection method, we design a lightweight CNN structure, which is shown in Table~\ref{tbCNNStru}. As shown in Table~\ref{tbCNNStru}, \emph{Layer1} and \emph{Layer4} are the convolutional layers whose filter size is $5\times5$ and the number of output feature maps is 16; \emph{Layer2} and \emph{Layer6} are the max-pooling layers whose filter size is $2\times2$; \emph{Layer3} and \emph{Layer5} are the local contrast normalization layers defined in Eq.~\ref{lcnEq}; \emph{Layer7} is a fully connected layer.   The hyperparameters used for training are given in Table~\ref{tbCNNPara}, where \emph{epsW} and \emph{epsB} respectively denote the learning rate for the weight vector and the bias in the CNN model; \emph{momW} and \emph{momB} respectively denote the momentum of the weight vector and that of the bias in the CNN model; $wc$ denotes the weight decay parameter. In the training process, we use 16 batches of images for training and 3 batches of images for validation. The whole training process takes around one hour on a computer equipped with a GTX-780Ti GPU. The trained CNN achieves 99.74\%  of test precision on the two validation batches. Given $H[w]=0.0008$, two CNN models (calculated by Eq.~\eqref{ensembleEq}) are trained, and the filters at the convolution layers in one model are shown in Fig.~\ref{convFilters}.
\begin{table}\small
\begin{center}
\caption {\label{tbCNNPara}The hyperparameters used for training the CNN model of the proposed method.}
\begin{tabular}{|l|c|c|c|c|c|}
  \hline
  LayerType & epsW & epsB & momW & momB & wc \\\hline
  Layer1 & 0.0030 & 0.0040 & 0.9000 & 0.9000 & 0.0000 \\\hline
  Layer4 & 0.0001 & 0.0002 & 0.9000 & 0.9000 & 0.0000\\\hline
  Layer7 & 0.0002 & 0.0003 & 0.9000 & 0.9000 & 0.0100 \\\hline
  LayerType & scale & pow & - & -& -\\\hline
  Layer3 & 0.0010 & 0.7500 & - & - & - \\\hline
  Layer5 & 0.0010 & 0.7500 & - & - & - \\
  \hline
\end{tabular}
\end{center}
\end{table}

\begin{figure*}[t]
\begin{center}
   \includegraphics[width=0.98\linewidth]{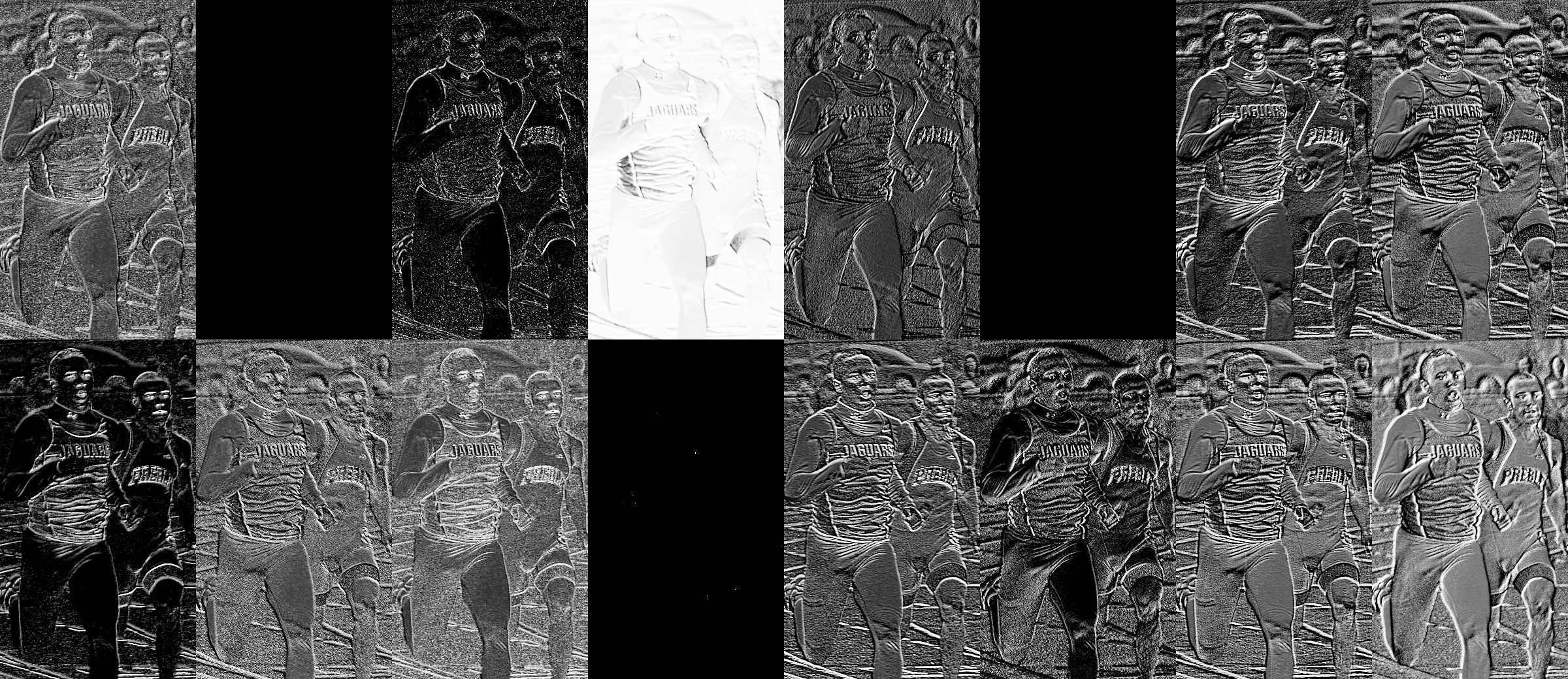}
\end{center}
   \caption{One fragment of DCFs at the sixth layer. The number of the feature maps in one fragment is equal to the number of channels at the sixth layer in Table 1.}
\label{oneFragment}
\end{figure*}
\textbf{Test environment:} For face detection, our method is programmed in Matconvnet~\cite{vedaldi15matconvnet} and runs on a computer equipped with a i7-4.0 GHZ CPU, a GTX-780Ti GPU and 32G RAM.
We evaluate the proposed face detection method on the FDDB~\cite{fddbTech} and the AFW~\cite{Ramanan} datasets. The FDDB dataset contains 5,171 faces in 2,845 images and the AFW dataset contains 468 faces in 205 images.
\subsection{Visualization of the trained CNN}
After training the parameters of the proposed method using the $cuda-convnet$ framework, the trained model is applied to detect the faces in the test images. One fragment of the DCFs is shown in Fig. \ref{oneFragment}. We can see that the intensities of most pixels in some feature maps are zero due to the usage of ReLU, and the sparse ratio is approximately equal to 0.5.

\begin{figure}
\begin{center}
   \includegraphics[width=0.98\linewidth]{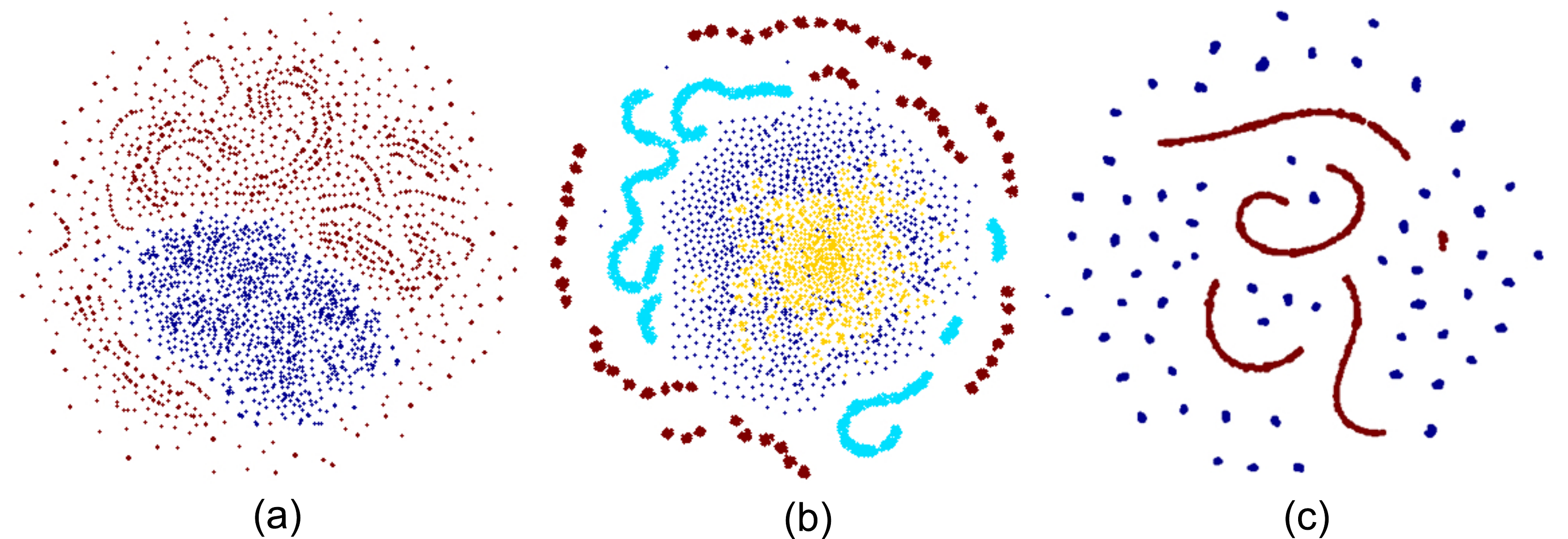}
   \caption{The visualization of the separability of the features extracted by CNN with different architectures. (a) General features of $32 \times 32$ images extracted by CNN ( the blue and red points denote the face features and the nonface features, respectively ). (b) The resized DCFs in accompany with the general features (also see Fig. \ref{framework} for details). (c) The features of $32 \times 32$ images extracted by CNN with the activation function of sigmoid ( the red and blue points denote the face features and the nonface features, respectively). }
\label{featVisu}
\end{center}
\end{figure}

In order to analyze the scale invariance of DCFs, multi-scale DCFs are obtained by resizing the DCFs from the original images using the nearest neighbor interpolation method. The multi-scale DCFs corresponding to the faces (represented by the bounding boxes) in the test images are resized to the size of the fully-connected layer. For fair  comparison, the general features before the fully-connected layer in the traditional CNN~\cite{yannCNN} and the dense features obtained by the proposed method with the sigmoid activation function are also evaluated. All the CNN models use the same structure as shown in Table 1. Using the t-SNE~\cite{maaten2008visualizing} visualization algorithm, 5,000 randomly selected resized DCFs and general features are shown in Fig.~\ref{featVisu}. From the figure, we can see that the resized DCFs are still robust for linear classification since the DCFs are sparse. However, the dense features extracted by the traditional CNN are not robust for linear classification.

\begin{table}\small
\begin{center}
{\renewcommand{\arraystretch}{1.5}
\caption {\label{tbFlops}The FLOPS[$\cdot10^{10}$] required by the three methods at three layers with a given parameter setting.}
\begin{tabular}{|l |c| c| c|| c|}
  \hline
 \multirow{2}{*}{}  &\multicolumn{3}{|c||}{Layer}  & \multirow{2}{*}{Total} \\  \cline{2-4}
  &1       & 4         & 7          &        \\ \cline{1-5}
  $A$       & 800$\times$ 600 & 800$\times$ 600  &800$\times$ 600    &       -      \\ \cline{1-5}
  $|P_{l-1}|$          & 1       & 16        & 16         &         -    \\ \cline{1-5}
  $|P_l|$              & 16      & 16        & 16         &         -    \\ \cline{1-5}
  $A_l$   & 800$\times$600 & 400$\times$300 & 200$\times$150    &     -        \\ \cline{1-5}
  $s_l$                & 25      & 25        & 1024       &       -      \\ \cline{1-5}
  $F_l$                & 1       & 4         & 16         &        -     \\ \hline
  $FLOPS^{Patch}$ & 34.56$\times$5 & 157.2864$\times$5& 0.98304$\times$5  & 964.1472    \\ \hline
  $FLOPS^{Image}$      & 0.002$\times$5 & 0.6144$\times$5  & 0.98304$\times$5  & 7.9972      \\ \hline
  $FLOPS^{DCF}$        &  0.002  & 0.135168  & 0.00096$\times$5  & 0.141968    \\ \hline
\end{tabular}}
\end{center}
\end{table}
\subsection{The Performance Comparison for Face Detection}
\begin{figure}
\begin{center}
 \includegraphics[width=0.78\linewidth ]{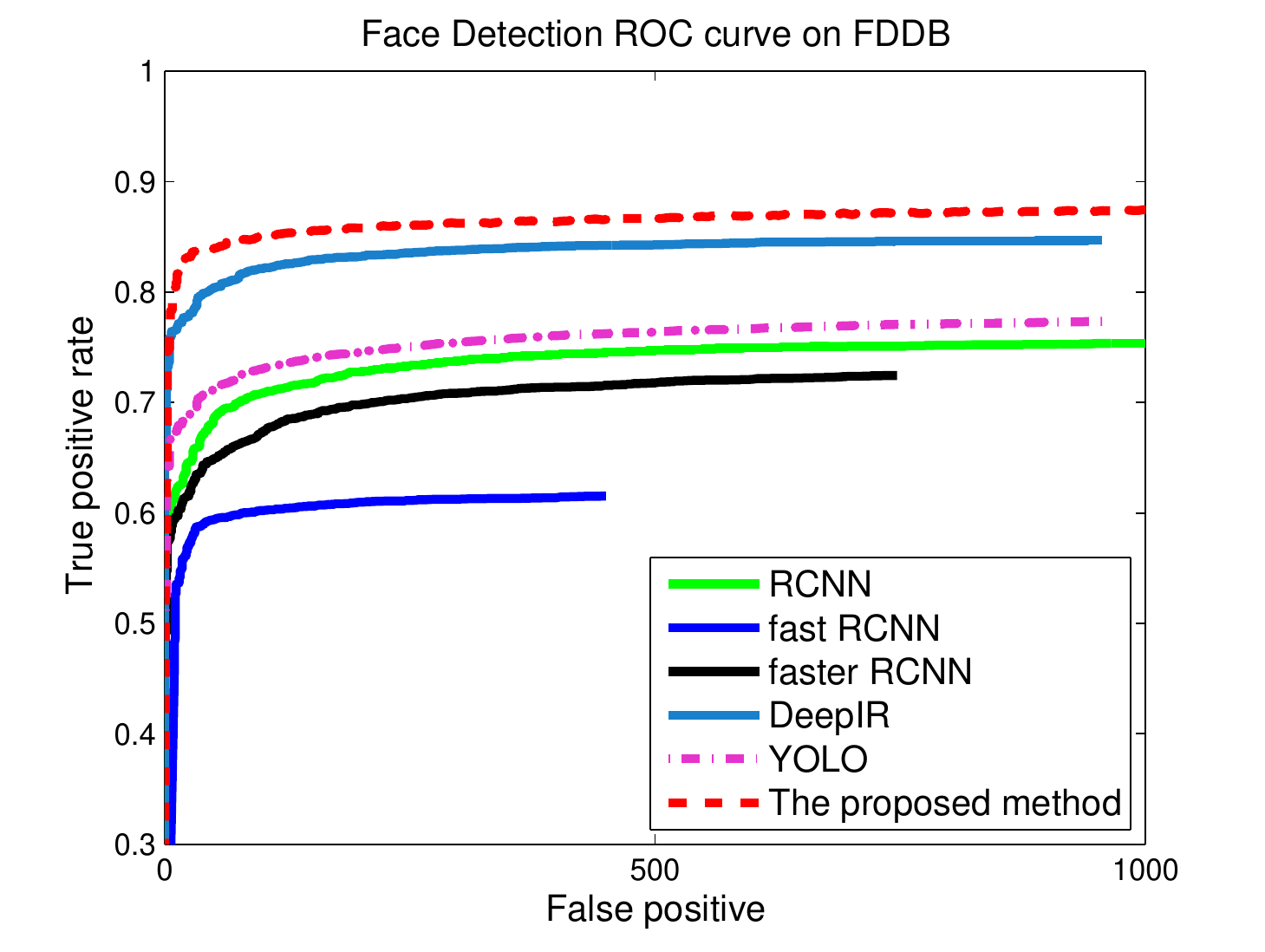}
\end{center}
 \vspace{-12pt}
   \caption{The comparison of the discrete ROC curves obtained by the six face detection methods based on CNN on the FDDB dataset.}
\label{FDDBROC_Deep}
\end{figure}

\begin{table}\small
\begin{center}
\caption {\label{tbTime}The CPU and GPU time (in seconds) used by the seven CNN based face detection methods and the proposed face detection method on a $800\times600$ test image. Note that YOLO is implemented in C/C++, and the other competing methods are implemented in Matlab and C/C++.}
\begin{tabular}{|l|c|c|c|}
  \hline
  Method & CPU time [s] & GPU time [s] &Total time\\\hline
  Patch-based & 2.3 & 25.1& 28.1 \\\hline
  Image-based & 0 & 2.7&2.7 \\\hline
  SPP-net & 2.3 & 0.3 &2.6 \\\hline
  Fast R-CNN & 2.7 & 10.1 &12.8      \\ \hline
  Faster R-CNN & 0.3 & 0.4 &0.7      \\ \hline
  DeepIR & 0.3 & 0.4 &0.7      \\ \hline
  YOLO & 0 & 0.04 &0.04      \\ \hline
  DCF-based & 0.1 & 0.09 &0.19      \\
  \hline
\end{tabular}
\end{center}
\end{table}
In this subsection, we compare the computational complexity between the proposed method and several other methods (including the image-based and patch-based methods) for face detection. We assume that five scales are used for detection and we use a test image with the size of 800$\times$600 for evaluation. The FLOPS required the three methods at different layers (i.e., 1, 4, 7) in CNN is shown in Table~\ref{tbFlops}. Moreover, Table~\ref{tbTime} gives the CPU and GPU time used by the four CNN-based face detection methods on a 800$\times$600 test image. In Table~\ref{tbTime}, for the patch-based method, we use the representative R-CNN, while we adopt the method of Giusti et al.~\cite{giustiICIP13} for the image-based method. In addition, SPP-net, Fast R-CNN, Faster R-CNN, DeepIR and YOLO are compared since these face detection methods are closely related to the proposed face detection method. SPP-net is similar to the proposed DCFs-based face detection method, because both methods directly perform detection on feature maps. However, compared with the proposed method,  SPP-net uses a more complicated CNN structure, which has more computational burden for face detection. From Tables~\ref{tbFlops} and~\ref{tbTime}, we can see that the proposed method is much more efficient than the patch-based method, the image-based method and the SPP-net method considering both theoretically computational complexity (i.e., the required FLOPS in Table~\ref{tbFlops}) and running time (in Table~\ref{tbTime}).  The proposed face detection method also runs faster than the fast R-CNN, faster R-CNN and DeepIR methods. The reason is that the CNN structures used in the three competing methods are more complex than that of the proposed method, which results in more computation. Moreover, the object proposal generation methods employed in the three competing methods consume more CPU time (about 2 seconds) than the proposed method (about 0.1 seconds). In term of GPU time, the running time of the proposed method is 0.09, which is faster than that of the other six competing methods (about 3.3 to 278.9 times faster) except for YOLO.  The YOLO method, which is implemented in C/C++, runs faster than the proposed face detection method, which is implemented in Matlab and C/C++. However, the proposed face detection method obtains higher true positive rate than YOLO (see Fig.~\ref{FDDBROC_Deep}).
\begin{figure}
\begin{center}
 \subfigure[]{\includegraphics[width=0.85\linewidth ]{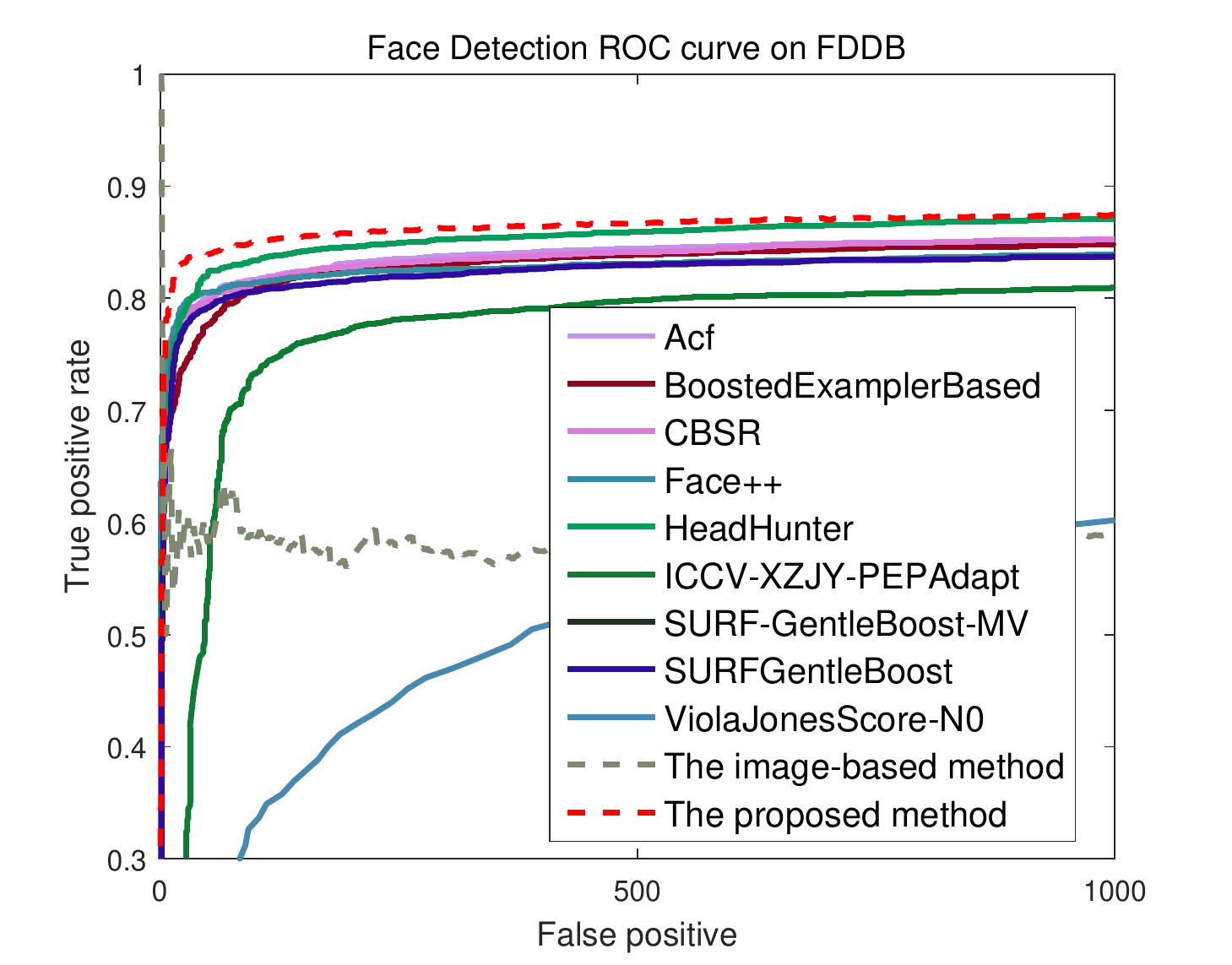}}
 \subfigure[]{\includegraphics[width=0.85\linewidth ]{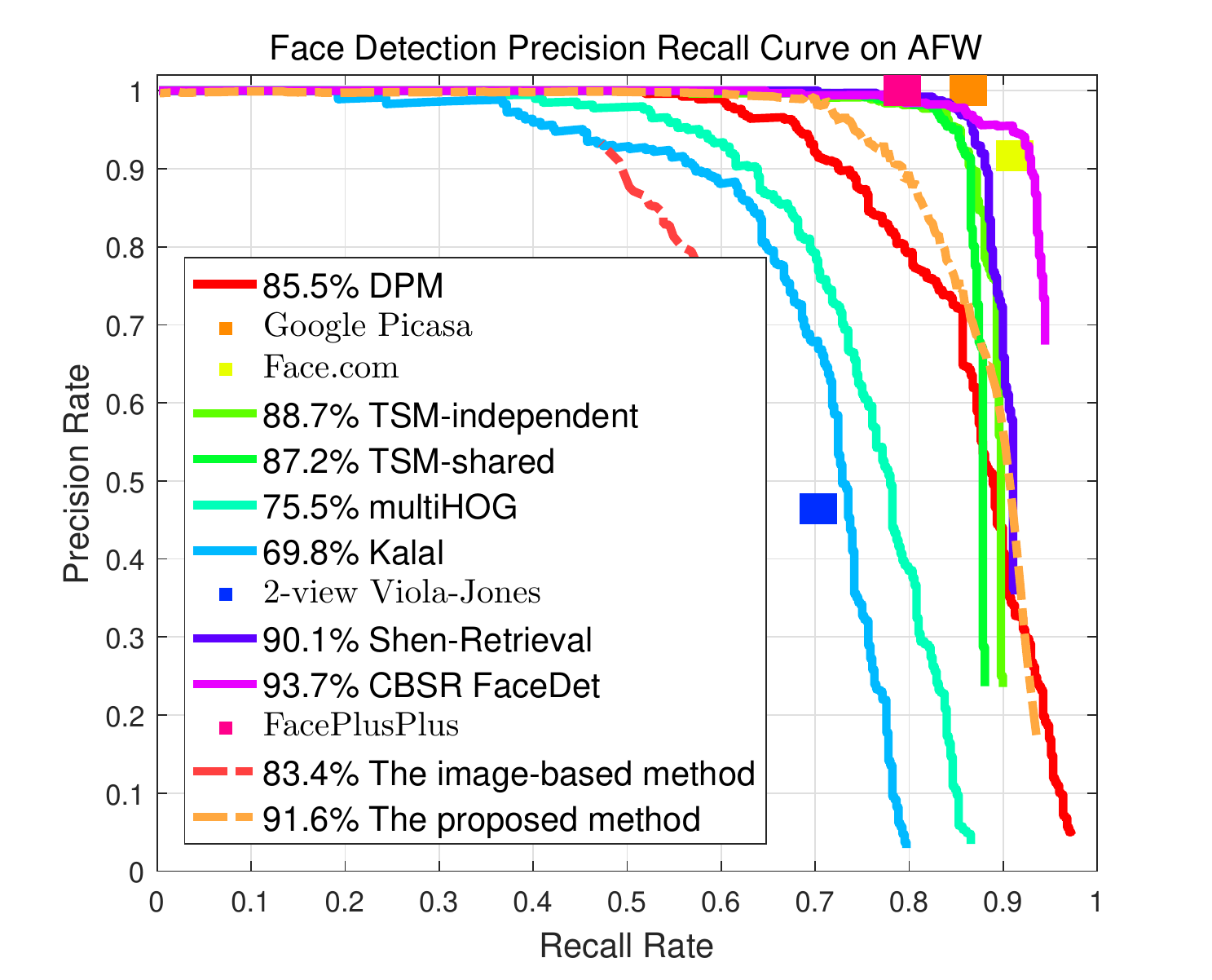}}
\end{center}
\vspace{-12pt}
   \caption{The comparison of the discrete ROC and PR curves for face detection obtained by the competing methods on the FDDB and AFW datasets. (a) The results of ROC curves obtained by the competing methods on the FDDB dataset. (b) The results of PR curves obtained by the competing methods on the AFW dataset.}
\label{FDDBROC}
\end{figure}

Next, we compare the performance obtained by different methods for face detection on the FDDB and AFW datasets. On the FDDB dataset, we use the criterion given in \cite{fddbTech,ZHAN201619}, which measures the degree of match between a detected and a manually annotated face region. The cutoff ratio of the intersected areas to the joined areas of the two regions is set to 0.5 in our case. We compare the proposed method with the following detection methods: (1) ACF-multiscale \cite{ACF-multiscale}, (2) HeadHunter \cite{HeadHunter}, (3) CBSR \cite{FDPM}, (4) Boosted Exemplar \cite{BoostedExemplar}, (5) SURF frontal/multiview \cite{SURF_frontal_multiview}, (6) SURF GentleBoost \cite{SURF_frontal_multiview}, (7) Face++ \cite{face++}, (8) PEP-Adapt \cite{PEP-Adapt}, (9) Viola-Jones \cite{Viola01robustreal-time} and (10) the image-based method \cite{giustiICIP13}. We select these methods since most of these methods are the state-of-the-art face detection methods based on hand-crafted features, and their results have been reported in the FDDB website~\cite{fddbTech}. In addition, we apply the image-based method to face detection as a comparison. The methods in~\cite{Li2015,zekun17}, which also use CNN for the task of face detection, are not compared since their source codes are not available in public. The comparison of the discrete ROC curves obtained by the competing methods is shown in Fig.~\ref{FDDBROC}(a), from which we can see that the proposed method achieves better performance than most of the competing methods except for HeadHunter. Due to the fact that the proposed method uses CNN to learn effective features of faces, the proposed method obtains better performance than that obtained by the most other competing methods. Although the image-based method also learns features for faces by using CNN, it obtains worse performance than that obtained by the other competing methods except for Viola-Jones. The main reason is that the features obtained the image-based method are less discriminative and not robust to scale variations. Moreover, the image-based method does not use an extra CNN model to eliminate false detection.

On the AFW dataset, we report the Precision-Recall curves for evaluation. Using the same settings as in \cite{FDPM}, we compare the proposed method with the following eleven methods: (1) DPM \cite{Ramanan}, (2) CBSR \cite{FDPM}, (3) TSM \cite{Ramanan}, (4) Kalal \cite{Face-TLD}, (5) Face++ \cite{face++}, (6) Google Picasa, (7) Face.com, (8) multiHOG \cite{Ramanan}, (9) Viola-Jones \cite{Viola01robustreal-time}, (10) Shen-Retrieval \cite{shen}, and (11) Image-based method \cite{giustiICIP13}. We select these methods since they are state-of-the-art face detection methods, which have been tested on the AFW dataset. The comparison of the Precision-Recall curves is shown in Fig.~\ref{FDDBROC}(b). The average precision obtained by the proposed method on the AFW dataset is 91.6\%, which is worse than CBSR, Google Picasa and Face.com due to the lack of a large number of training faces in the wild. However, the proposed method shows excellent classification performance when the recall ratio is high. The recall ratio obtained by the proposed method is higher than that obtained by DPM, Kalal, multiHOG, Viola-Jones and the image-based method when the precision ratio is greater than 0.7.

Several object detection methods based on CNN (such as R-CNN, faster R-CNN, YOLO) can be used for face detection, and they can achieve comparative performance on the FDDB dataset. To show the effectiveness of the proposed method, we implement and evaluate several state-of-the-art object detection methods on the FDDB dataset. We compare the proposed method against the following five face detection methods based on CNN: (1) R-CNN~\cite{jiang2016}, (2) fast R-CNN~\cite{Triantafyllidou2017}, (3) faster R-CNN~\cite{jiang2016}, (4) DeepIR~\cite{SunWH17} and (5) YOLO~\cite{yolo2016}. The five competing methods perform face detection by using CNN classifiers on the face proposals generated by different object proposal generation methods. The YOLO method applies a single neural network to a whole image. The network divides the image into regions and predicts bounding boxes and probabilities of each region being an object. Then, these bounding boxes are weighed by the predicted probabilities. Fig.~\ref{FDDBROC_Deep} shows the ROC curves obtained by the six competing methods on the FDDB dataset. As can be seen, the proposed method obtains the best performance. The reason is that the proposed method can obtain effective features when a sliding window strategy is applied on DCFs. The R-CNN, fast R-CNN and faster R-CNN methods obtain worse performance than the other competing methods since the performance of the three methods for face detection is greatly affected by the quality of object proposals generated  by the object proposal generation methods. More specifically, both R-CNN and fast R-CNN use the Selective Search object proposal generation method (SS) to generate object proposals, and faster R-CNN uses the Region Proposal Network (RPN) to generate object proposals. However, both SS and RPN are not effective at generating object proposals for faces, especially for small-sized faces. DeepIR improves RPN  by re-training the PRN model on the face dataset, and it increases the discriminative ability of the classifier by using the hard negative mining strategy. Thus, DeepIR obtains better performance than R-CNN, fast R-CNN and faster R-CNN. However, DeepIR obtains worse performance than the proposed method since the latter directly uses a sliding window strategy on the DCFs, which can achieve high face recall than the object proposal generation method used in DeepIR.

\begin{figure*}
\begin{center}
 \includegraphics[width=0.85\linewidth ]{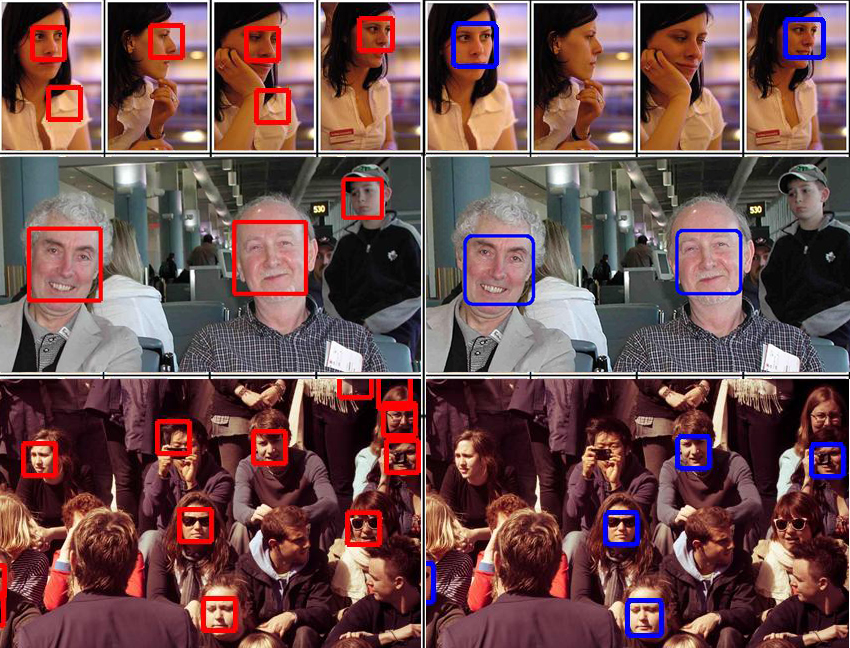}
\end{center}
   \caption{The comparison results of face detection. Red bounding boxes (left) denote the result obtained by the proposed method. Blue bounding boxes (right) denote the result obtained by Face++.}
\label{compRe}
\end{figure*}
Some detection results are given in Fig. \ref{compRe}. We can see that the proposed method can find more correct faces with occlusions and different poses, where the false detection can be eliminated by using an extra CNN model during the classification process.

\section{Conclusion}
In this paper, we propose a fast face detection method based on DCFs extracted by an elaborately designed CNN. Compared with the state-of-the-art face detection methods using CNN, the proposed method performs direct classification on DCFs, which can significantly improve the efficiency during the face detection process. Moreover, the proposed method can effectively detect small-sized faces by using the sliding window strategy on DCFs. In contrast, current state-of-the-art face detection methods based on CNN are hard to detect small-sized faces since the performance of these methods is greatly affected by the quality of object proposals generated by the object proposal generation methods.  Experimental results have shown that the proposed DCFs-based face detection method can achieve promising performance on several popular face detection datasets.

\section*{Acknowledgments}

This work was supported by the National Natural Science Foundation of China under Grants U1605252, 61472334, 61571379 and 61370124, and by the Natural Science Foundation of Fujian Province of China under Grant 2017J01127.

\section*{References}

\bibliography{detectionbib}

\end{document}